\documentclass{svproc}
\usepackage{url}
\usepackage[linesnumbered, ruled, vlined]{algorithm2e}
\usepackage{array,multirow}
\usepackage{graphicx}

\begin{document}
\mainmatter              
\title{eLIAN: Enhanced Algorithm for Angle-constrained Path Finding}
\titlerunning{eLIAN: Enhanced Algorithm for Angle-constrained Path Finding}  
%
\author{Anton Andreychuk\inst{1} \and Natalia Soboleva\inst{2}
\and Konstantin Yakovlev\inst{2, 3, 4}}
\authorrunning{Anton Andreychuk et al.} 
%
\tocauthor{Anton Andreychuk, Natalia Soboleva, Konstantin Yakovlev}
\institute{Peoples' Friendship University of Russia, Moscow, Russia \\
\email{andreychuk@mail.com}
\and
National Research University Higher School of Economics, Moscow, Russia \\
\email{nasoboleva@edu.hse.ru, kyakovlev@hse.ru}
\and
Federal Research Center ``Computer Science and Control'' of Russian Academy of Sciences, Moscow, Russia \\
\email{yakovlev@isa.ru}
\and
Moscow Institute of Physics and Technology \\
\email{yakovlev.ks@mipt.ru}}

\maketitle
\begin{abstract}
Problem of finding 2D paths of special shape, e.g. paths comprised of line segments having the property that the angle between any two consecutive segments does not exceed the predefined threshold, is considered in the paper. This problem is harder to solve than the one when shortest paths of any shape are sought, since the planer's search space is substantially bigger as multiple search nodes corresponding to the same location need to be considered. One way to reduce the search effort is to fix the length of the path's segment and to prune the nodes that violate the imposed constraint. This leads to incompleteness and to the sensitivity of the 's performance to chosen parameter value. In this work we introduce a novel technique that reduces this sensitivity by automatically adjusting the length of the path's segment on-the-fly, e.g. during the search. Embedding this technique into the known grid-based angle-constrained path finding algorithm – LIAN, leads to notable increase of the planner's effectiveness, e.g. success rate, while keeping efficiency, e.g. runtime, overhead at reasonable level. Experimental evaluation shows that LIAN with the suggested enhancements, dubbed eLIAN, solves up to 20\% of tasks more compared to the predecessor. Meanwhile, the solution quality of eLIAN is nearly the same as the one of LIAN.
\end{abstract}

\section{Introduction}
Path finding is a vital capability of any intelligent agent operating in real or simulated physical environment. In case this environment is known a priory and remains static, grids are commonly used to explicitly represent it as they are simple yet informative and easy-to-update graph models. Grids appear naturally in game development \cite{sturtevant2012benchmarks} and are widely used in robotics \cite{elfes1989using}, \cite{thrun2003learning}. When agent's environment is represented by a grid, heuristic search algorithms, such as A* \cite{hart1968formal} and its modifications, are typically used for path planning. The path planning process is carried out in the state-space, where states are induced by grid elements, and the function of generating successors defines which movements on the grid are allowed or not \cite{yap2002grid}.

\begin{figure}[b!]
\includegraphics[width=\textwidth]{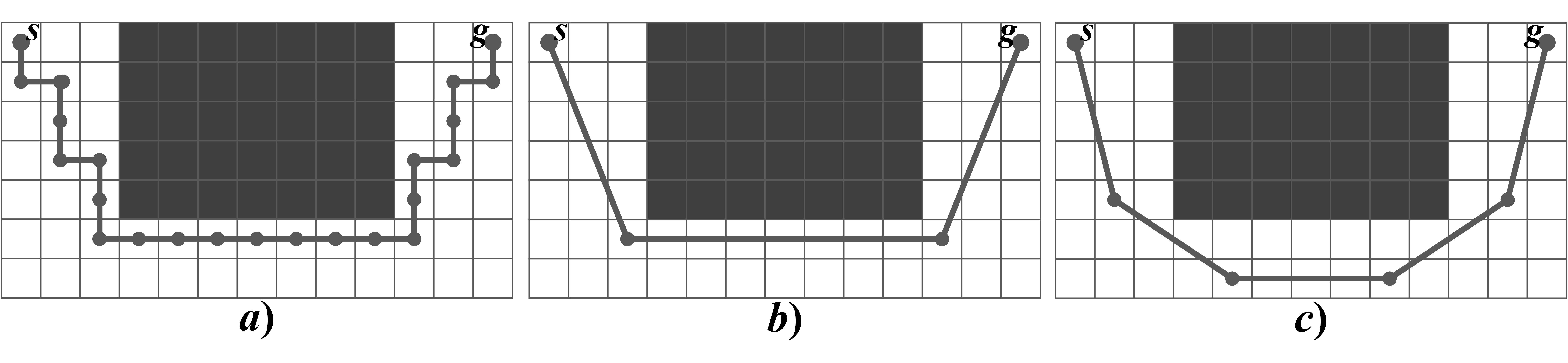}
\caption{Different types of grid paths. a) A*-path composed of the moves between cardinally adjacent cells; b) any-angle path; c) angle-constrained path}
\label{fig1}
\end{figure}

Numerous works on grid-based path finding allow agent to move only between the cardinally adjacent grid elements, e.g cells or corners (sometimes, diagonal moves are allowed as well), see \cite{harabor2011online}, \cite{botea2004near}, \cite{silver2005} for example. This often results in finding paths, containing multiple sharp turns (see \figurename \ \ref{fig1}a), which might be undesirable for some applications. To mitigate this issue a few techniques have been proposed. First, smoothing can be carried out as the post-processing step \cite{botea2004near}, i.e. when the path has already been planned. Second, smoothing can be interleaved with the state-space exploration resulting in so-called any-angle behavior (see \figurename \ \ref{fig1}b). Originally any-angle algorithms, like Theta* \cite{daniel2010theta}, Lazy Theta* \cite{nash2010lazy} were lacking optimality, by recently ANYA* \cite{harabor2016} was introduced which guarantees finding shortest any-angle paths between the given grid cells. 

Any-angle algorithms allow agent to move in any arbitrary direction, but the resulting paths still might contain sharp turns (see \figurename \ \ref{fig1}b). \cite{xu2013planning} addresses this issue by keeping the number of turns to a possible minimum. In \cite{kim2014angular} a modification of Theta* was proposed, that prunes away the nodes that lead to the turns violating the given kinematic constraints of the agent. Finally, in \cite{yakovlev2015grid} LIAN algorithm was proposed that seeks for the so-called angle-constrained paths, e.g. the paths comprised of line segments (with the endpoints tied to the grid elements), having the property that the angle between any two consecutive segments does not exceed the predefined threshold (see \figurename \ \ref{fig1}c). In this work we focus on these sort of paths, e.g. angle-constrained paths.

Finding angle-constrained paths is a much harder problem to solve compared to finding shortest paths, as the algorithm has to keep numerous nodes corresponding to the same grid element because the latter can be approached from different directions, rather than keeping only one node and re-writing its g-value (cost of the best path known so far). In order to keep the number of nodes in the search space to a reasonable minimum LIAN considers only states that correspond to the moves of the fixed length, which is the parameter of the algorithm. This, obviously, leads to incompleteness and to the sensitivity of the planner's performance to the chosen parameter value. In this work, we mitigate this issue by introducing an original technique that adjusts the parameter value on-the-fly while the algorithm performs the search. This does not make the algorithm complete but significantly increases the chance of finding the solution. As shown experimentally, success rate of the enhanced planner, dubbed eLIAN (enhanced LIAN) exceeds the one of the original algorithm up to 20\%. Meanwhile, the solution quality of eLIAN and LIAN is nearly the same.

\section{Background}
\subsubsection{Problem statement}
Consider a grid composed of blocked and unblocked cells and a point-agent that is allowed to move from one unblocked cell to the other following the straight line segment connecting the centers of those cells. The move is considered feasible if there exists line-of-sight between the endpoints of the move, in other words -- if the corresponding segment of the straight line does not intersect any blocked cell. In practice the well-known in computer graphics  algorithm from \cite{bresenham1965algorithm} adopting slight modifications can be used to efficiently perform line-of-sight checks. 

Consider now a path, $\pi$, composed of feasible moves, $e_i$, and the value $\alpha_m(\pi) = max (|\alpha(e_1, e_2)|, |\alpha(e_2, e_3)|, \ldots, |\alpha(e_{v-1}, e_v)|)$, where $\alpha(e_j, e_{j+1})$ is the angle between two consecutive line segments representing the moves. The task of the planner is formalized as follows. Given the start and the goal cell, as well as the constraint $\alpha_{MAX} \in [0^\circ; 180^\circ]$, find a feasible path $\pi$, connecting them, such that $\alpha_m(\pi) \leq \alpha_{MAX}$.

\subsubsection{LIAN algorithm}
LIAN (abbreviation for "limited angle") is a heuristic search algorithm that seeks for the angle-constrained path composed of the segments of the fixed length, $\Delta$, which is the input parameter of the algorithm.

LIAN search node $\textbf{s}$ is identified by the tuple $[\textit{s}, \textbf{\textit{bp}}(\textbf{s})]$, where $\textit{s}$ is the grid cell and $\textbf{\textit{bp}}(\textbf{s})$ is the back-pointer to the predecessor of $\textbf{s}$, e.g. the node which was used to generate $\textbf{s}$. The back-pointer to the predecessor allows to identify the heading, i.e. the direction of movement of the agent in case if it arrives to $s$ from \textbf{\textit{bp}}(\textbf{s}). Additionally such data as the \textit{g}-value -- the length of the angle-constrained path from the start node to the current node and the \textit{h}-value -- approximate length of the path to the goal is also associated with the node.

\begin{figure}[t!]
    \centering
    \includegraphics[width=0.9\textwidth]{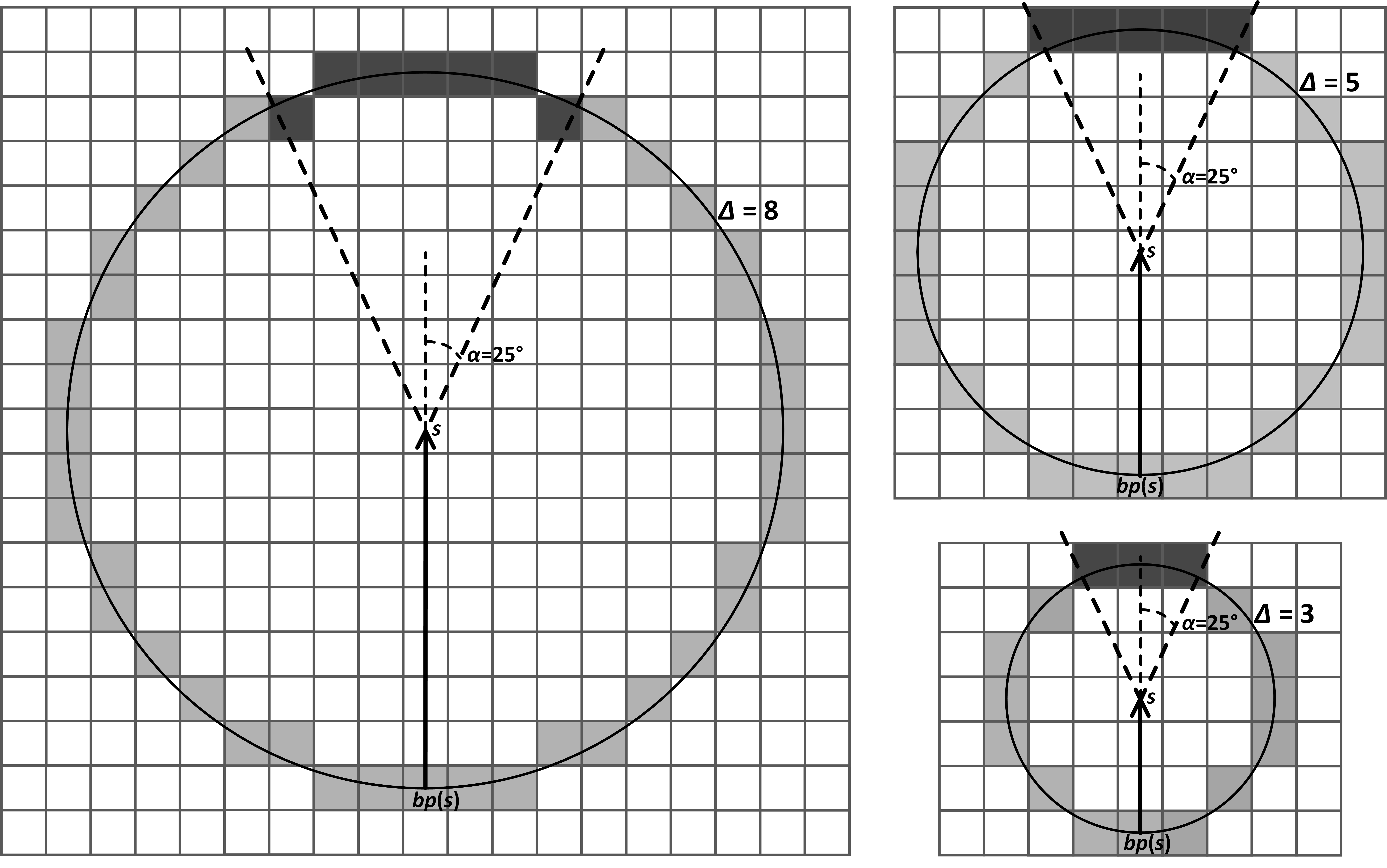}
    \caption{The process of generating successors. The cells identified by midpoint circle algorithm are marked by light-gray. Among them only those cells are considered that satisfy the angle constraint (marked by dark-gray).}
    \label{fig1add}
\end{figure}

LIAN explores the search-space in the same way A* does. On each step it chooses the most promising node, e.g. the one minimizing \textit{f}-value ($f(\textbf{\textit{s}})=g(\textbf{\textit{s}})+h(\textbf{\textit{s}})$) from \textit{OPEN} (set of previously generated nodes which are the candidates for further processing) and expands it. Expanding includes removing the node from \textit{OPEN}, adding it to \textit{CLOSED} (set of already processed nodes), generating valid successors and adding them to \textit{OPEN}. Generating successors is a multi-step procedure including: \textit{a}) estimating the cells that lie within $\Delta$-distance, \textit{b}) pruning away the cells that result in a move violating given angle constraint; \textit{c}) pruning away the cells that violate line-of-sight constraint; \textit{d}) pruning away the cells that result in generating nodes that have already been visited before. To find the cells lying within $\Delta$-distance midpoint circle algorithm is used that builds discrete approximation of a circumference centered in a given cell and of a given $\Delta$-radius. Such circumferences are depicted on \figurename \ \ref{fig1add}. As one can see, the number of cells that satisfy angle-constraints grows with $\Delta$.

Algorithm stops when either the goal node is retrieved from \textit{OPEN} or the \textit{OPEN} is exhausted. In the first case the sought angle-constrained path can be re-constructed using the back-pointers, in the second one algorithm returns failure to find the path.

For the sake of space LIAN pseudocode is omitted but one can consult eLIAN code given in Algorithm 1 and Algorithm 2. LIAN code is the same stating that $\Delta_{max} = \Delta_{min} = \Delta$ for LIAN. Thus, when expanding a node, condition on line 11 is always false for LIAN and lines 12-13 are skipped. Condition on line 17 is also always false thus line 18 is skipped as well.

\subsubsection{Sensitivity to the input parameter}
As mentioned before, in general LIAN provides no guarantees to find a path even though it exists. One can only claim that if the angle-constrained path comprised of $\Delta$-segments exists, the algorithm will find it \cite{yakovlev2015grid}. In practice quite often LIAN fails to find a solution due to inappropriately chosen $\Delta$-value.

\begin{figure}[b!]
\includegraphics[width=\textwidth]{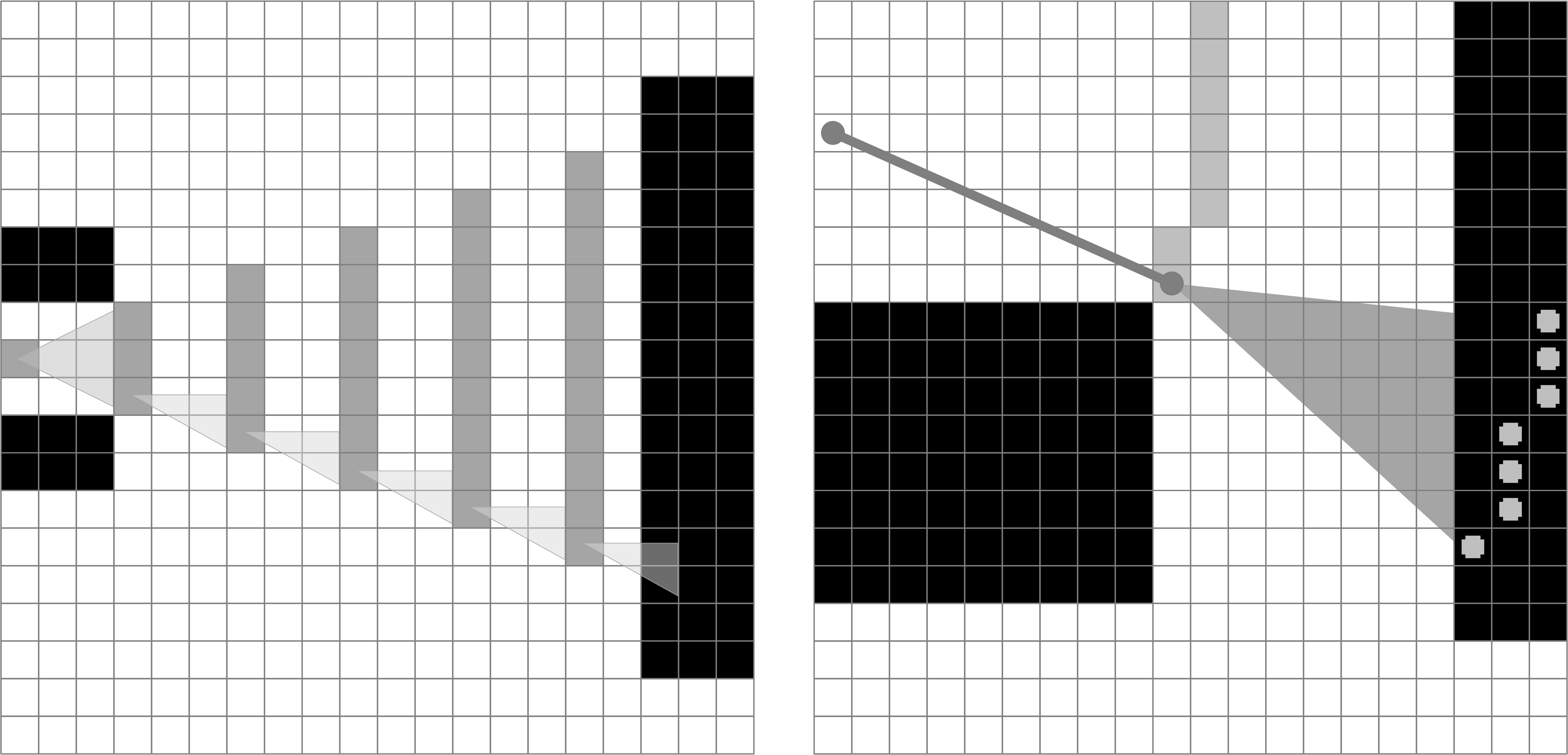}
\caption{Inappropriate $\Delta$-value for LIAN. Left:  $\Delta$-value is set too low, thus the number of successors on each step is limited, thus the obstacle can not be circumnavigated. Right:  $\Delta$-value is set too high, line-of-sight constraint is consequently breached as a result no valid successors are generated to continue the search.}
\label{fig2}
\end{figure}

On \figurename \ \ref{fig2} two scenarios are depicted when setting the $\Delta$-value too low/too high leads to failure. When $\Delta$ is low the number of cells lying within $\Delta$-distance is also low, thus the number of potential successors that satisfies the predefined angle constraint for each expansion is low. This can lead to the exhaustion of \textit{OPEN}, like in case depicted on \figurename \ \ref{fig2} left (here $\Delta=3$ and angle constraint is set to be $25^{\circ}$). Setting $\Delta$ to high values can also result in lowering down the number of successors but now it is mainly due to the violation of line-of-sight constraints in certain cases -- see \figurename \ \ref{fig2} right.

\section{eLIAN}
To mitigate the influence of the parameter $\Delta$ on the effectiveness of the algorithm we suggest adjusting the former on-the-fly thus adapting LIAN behavior to given path finding instance, e.g. to the configuration of the obstacles and to the particular locations of start and goal.

In both cases described above and shown on \figurename \ \ref{fig2} algorithm would not have failed to find the solution if $\Delta$ was either higher (the first case), or lower (the second case). Thus rather than fixing $\Delta$-value we suggest setting the upper and the lower limit, $\Delta_{max}$ and $\Delta_{min}$, and allow $\Delta$ to take values from the range $[\Delta_{min}, \Delta_{max}]$. 

\begin{algorithm}[t]
\caption{eLIAN--Main-Loop}
{
\textbf{\textit{bp}(start)}:=$\emptyset$; \textit{g}(\textbf{start}):=0\;
$\Delta(\textbf{start}):=\Delta_{max}$\;
$OPEN$.push(\textbf{start}); $CLOSED$:=$\emptyset$\;
\While{$OPEN$ is not empty}
{
\textbf{s}:=argmin$_{\textbf{s}\in OPEN}f(\textbf{s})$\;
$OPEN$.remove(\textbf{s})\;
\If{$s=goal$}
{
\textbf{return}
$getPathFromParentPointers(\textbf{s})$\;
}
$CLOSED$.push(\textbf{s})\;
$Expand(\textbf{s},\alpha_m, \Delta_{min})$\;
}
\textbf{return} ``path not found''\;
}
\end{algorithm}

Introducing the $\Delta$ range leads to the following questions: a) what is the initial value of $\Delta$ from this range; b) what is the procedure of adjusting $\Delta$; c) how is it embedded to the search algorithm. We suggest the following answers to these questions.

The initial value equals $\Delta_{max}$. The rationale behind this is that setting $\Delta$ to high values potentially leads to large number of successors, except the cases (discussed above), which are to be handled by the next adjusting procedure.

In case no successors are generated at some step when $\Delta_{max}$ is in use it should be lowered down by multiplying it by a factor of $k \in (0; 1)$, which becomes the algorithm's parameter. Now, in order to keep track of $\Delta$-values they should be associated with the search nodes like \textit{g}-values and \textit{h}-values. We will refer to the exact $\Delta$-value of the node as to $\Delta(\textbf{s})$. After fixing $\Delta(\textbf{s})$ for the current node $\textbf{s}$, the latter is inserted back to \textit{OPEN}, thus in case of re-expansion another set of successors will be generated for the node and possibly this set will not be empty. In case it is, $\Delta(\textbf{s})$ is decremented again. Lowering down the $\Delta$-value continues up to a point when $k \cdot \Delta(\textbf{s}) < \Delta_{min}$. If this condition holds, e.g. the $\Delta$-value can not be lowered down any more as it goes out of the range, the search node is discarded for good.

To provide additional flexibility to the algorithm we suggest not only decrementing $\Delta$-value but also incrementing it after a few consecutive successful attempts of successors' generation. Number of such attempts can be seen as another input parameter. As a result the algorithm tries to keep $\Delta$-value as high as possible and lowers it down only when needed.

\begin{figure}[b]
    \centering
    \includegraphics[width=0.9\textwidth]{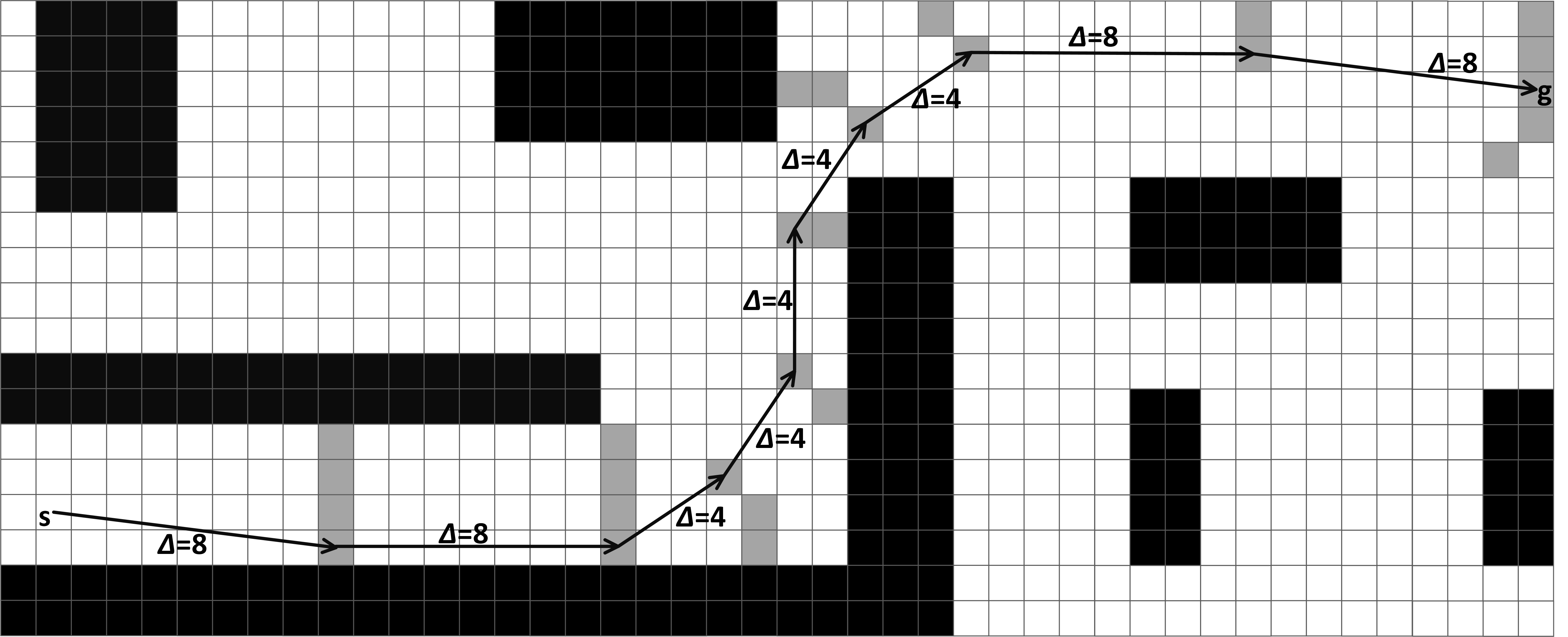}
    \caption{An example of how eLIAN works. Possible $\Delta$-values are 8 and 4. Angle constraint is $\alpha=35^\circ$}
    \label{figeLIAN}
\end{figure}

We name the resultant algorithm as eLIAN (stands for "enhanced LIAN"). Pseudocode of eLIAN is presented as Algorithm 1 (main loop) and Algorithm 2 (node expansion). Please note that the code increments $\Delta$-value after 2 consecutive successful expansions (see line 17 of the Algorithm 2), but it can be easily modified for another threshold. 

An example of eLIAN trace is presented in \figurename \ \ref{figeLIAN}. The initial $\Delta$-value is 8 and it is too high for the given task due to rather small passages between the obstacles. Regular LIAN algorithm would have failed to solve this task as it would not be able to generate valid successors that bypass the obstacles. In contrast, eLIAN decreased the $\Delta$-value to 4 after two steps and successfully found a sequence of small sections going through the passage. After the obstacle was successfully bypassed $\Delta$-value was increased back to 8 leading to more straightforward and rapid movement towards the goal.

\begin{algorithm}[t]
\caption{Expand(\textbf{s}, $\alpha_m$, $\Delta_{min}$)}
{
	$SUCC:=getDeltaSuccessors$(\textbf{s},$\Delta$(\textbf{s}))\;
	\If{\textup{$dist(s,goal)<\Delta$(\textbf{s})}}
	{
		$SUCC$.push(\textbf{goal})\;
	}
	\For{\textup{each \textbf{s$'$} $\in$ $SUCC$}}
	{
		\If{\textup{$lineOfSight$($s,s'$)=false \textbf{or} \newline $angleConstraint$($bp$($s$),$s$,$s'$)=false}}
		{
		$SUCC$.remove(\textbf{s$'$})\;
		}
		\For{\textup{each \textbf{s$''$} $\in CLOSED$}}
		{
			\If{\textup{$s'=s''$ \textbf{and} $bp$($s'$)=$bp$($s''$)}}
			{
				$SUCC$.remove(\textbf{s$'$})\;
			}
		}
	}
	\If{$SUCC=\emptyset$}
	{
		\If{\textup{$\Delta$(\textbf{s}) $>\Delta_{min}$}}
		{
			$\Delta$(\textbf{s}):= $k \cdot$  $\Delta$(\textbf{s}); //$k \in (0;1)$ is a parameter\\
			$OPEN$.push(\textbf{s})\;
		}
	}
	\Else
	{
		\For{\textup{each \textbf{s$'\in SUCC$}}}
		{
			\textit{g}(\textbf{s$'$}):=\textit{g}(\textbf{s})$+$\textit{dist}($s,s'$)\;
			\If{\textup{$\Delta$(\textbf{s})=$\Delta$(\textbf{\textit{bp}}(\textbf{s})) \textbf{and} $\Delta$(\textbf{s})$<$$\Delta_{max}$}}
			{
				$\Delta$(\textbf{s}$'$):=$\Delta$(\textbf{s})$/k$\;
			}
			$OPEN$.push(\textbf{s$'$})\;
		}
	}
}
\end{algorithm}

The following properties of eLIAN can be inferred.

\textit{Property 1}. eLIAN is correct, e.g. it terminates after the finite number of steps.

\textit{Sketch of proof}. Algorithm is performing the search until the \textit{OPEN} list is empty (or until the goal node is retrieved from it). \textit{OPEN} contains only elements that correspond to the grid cells the total number of which is finite. As the algorithm operates only by $\Delta$-sections the number of potential parents of the cell per each $\Delta$ is also finite. eLIAN operates finite number of $\Delta$-values in range $[\Delta_{min};\Delta_{max}]$. Thus the number of all nodes possibly considered by eLIAN is finite.

At the same time when a new node is generated the algorithm checks whether this node (the node defined by the same cell and the same parent) has been processed before already (lines 7-9 of Algorithm 2). And in case the \textit{CLOSED} already contains such node it is pruned and is not added to \textit{OPEN}. Taking into account the fact that on each step an element is removed from \textit{OPEN} (line 6 of Algorithm 1) and added to \textit{CLOSED} one can infer that sooner or later either the goal node will be retrieved or there will not be any node that could be added to \textit{OPEN} and it will become empty. In both cases (lines 7, 11 of Algorithm 1) algorithm terminates.

\textit{Property 2}. If there exists an angle-constrained path comprised of $\Delta_{max}$-segments, eLIAN will find a solution.

\textit{Sketch of proof}. As the path composed of $\Delta_{max}$-segments exists, eLIAN will always maintain at least one node lying on that path as the first expansion happens with $\Delta_{max}$. Thus eLIAN always has an option to expand this node and to generate another one lying on $\Delta_{max}$-path. Thus, sooner or later either this path will be found or eLIAN will find another path comprised of the segments of alternating lengths (due to the expansions of nodes with different $\Delta$-values that have lower \textit{f}-values).

\section{Experimental evaluation}
\begin{figure}[ht]
    \centering
    \includegraphics[width=0.99\textwidth]{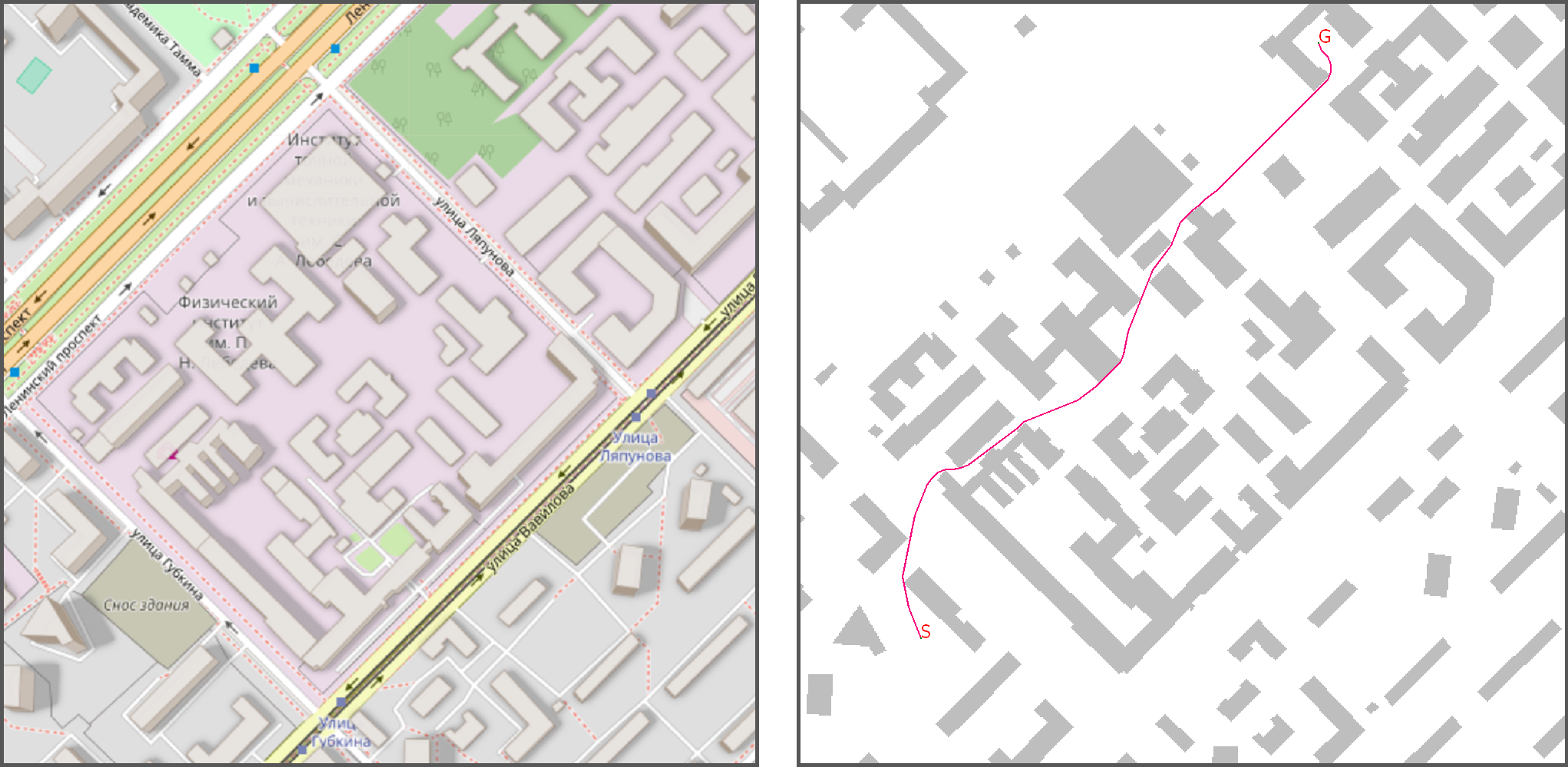}
    \caption{Left: Map fragment of an urban environment retrieved from OpenStreetMaps. Right: An example of a task, solved by eLIAN on the corresponding grid.}
    \label{figTaskExample}
\end{figure}

To empirically evaluate the suggested algorithm, eLIAN, and to compare it with the predecessor, LIAN, we used the well-known in the community benchmark sets from N. Sturtevant’s collection \cite{sturtevant2012benchmarks}: Baldur’s Gate (\textit{BG}) and Warcraft III (\textit{WIII}). Both collections contain maps used in computer games. \textit{BG} maps (75 in total) represent mostly indoor environments while \textit{WIII} maps (36 in total) -- imaginary outdoor environments. Benchmark repository contains path finding instances organized into the so-called buckets. Instances belonging to the bucket have nearly the same solution cost (path length as found by A*). We used instances that belong to the most "tough" buckets, e.g. the instances with the highest A* solution costs. There were taken 14 tasks per map for BG and 30 tasks for WIII to get more than 1000 tasks per each collection. We have also used a collection of maps that represent urban outdoor environment. These maps were retrieved from OpenStreetMaps\footnote[1]{http://www.openstreetmaps.org/export}. 100 maps of 1.2x1.2km fragments were converted to 501x501 grids (thus each grid cells corresponds to 2.7x2.7 area). 10 tasks per map were generated in such a way that a distance between start and goal is greater or equal to 1km (400 cells). An example of a map fragment that was used for the experimental evaluation and the corresponding grid are presented in \figurename \ \ref{figTaskExample}. 

Experimental evaluation was carried out on Windows operated PC with Intel Core 2 Duo Q8300 @2.5Ghz processor and 2GB of RAM. For the sake of fair comparison both algorithms were coded from scratch using the same programming techniques and data structures.

It is known from previous research that finding angle-constrained paths might take time, so inflated heuristic was used to speed-up the search. Heuristic weight was set to 2. Besides, 5 minutes time limit was introduced. In case the algorithm did not come up with the solution within this time, the result of the run was considered to be failure.

The following algorithms were evaluated: LIAN-20, e.g. the original algorithm with $\Delta=20$; eLIAN-20-10, e.g. $\Delta_{max}=20$, $\Delta_{min}=10$; eLIAN-20-5, e.g. $\Delta_{max}=20$, $\Delta_{min}=5$. For eLIAN we used $k=0.5$, e.g. $\Delta$-value was cut in half in case no valid successors were generated. This means that $\Delta$-value alternated between 10 and 20 for eLIAN-20-10 and between 5 and 10 and 20 for eLIAN-20-5. The angle constraint was set to be $20^{\circ}$, $25^{\circ}$ and $30^{\circ}$.

\begin{table}
\centering
\caption{Success rate of LIAN and eLIAN}
\label{tab1}
\resizebox{0.99\textwidth}{!}{\begin{tabular}{|c|ccc|ccc|ccc|}
\hline
&\multicolumn{3}{|c|}{City Maps}&\multicolumn{3}{|c|}{Baldur's Gate}&\multicolumn{3}{|c|}{Warcraft III}\\
&$20^\circ$&$25^\circ$&$30^\circ$&$20^\circ$&$25^\circ$&$30^\circ$&$20^\circ$&$25^\circ$&$30^\circ$\\
\hline
LIAN-20&84.2\%&89.0\%&91.8\%&61.24\%&71.71\%&81.05\%&78.33\%&82.22\%&85.74\%\\
\multirow{2}{*}{eLIAN-20-10}& 89.5\%& 93.5\%& 95.2\%& 74.38\%& 84.76\%& 88.0\%& 83.98\%& 88.33\%& 91.76\%\\
&(+5.3)&(+4.5)&(+3.4)&(+13.14)&(+13.05)&(+6.95)&(+5.65)&(+6.11)&(+6.02)\\
\multirow{2}{*}{eLIAN-20-5}&92.4\%&95.4\%&97.0\%&82.38\%&88.29\%&87.52\%&87.5\%&92.59\%&93.61\%\\
&(+8.2)&(+6.4)&(+5.2)&(+21.14)&(+16.57)&(+6.48)&(+9.17)&(+10.37)&(+7.87)\\
\hline
\end{tabular}}
\end{table}

\begin{table}
\centering
\caption{Normalized number of tasks that were not solved by LIAN but solved with eLIAN}
\label{tab2}
\resizebox{0.99\textwidth}{!}{\begin{tabular}{|c|ccc|ccc|ccc|}
\hline
&\multicolumn{3}{|c|}{City Maps}&\multicolumn{3}{|c|}{Baldur's Gate}&\multicolumn{3}{|c|}{Warcraft III}\\
&$20^\circ$&$25^\circ$&$30^\circ$&$20^\circ$&$25^\circ$&$30^\circ$&$20^\circ$&$25^\circ$&$30^\circ$\\
\hline
eLIAN-20-10& 27.8\%&35\%&37.2\%& 34.7\%& 49.1\%& 44.7\%& 17.5\%& 22.6\%& 32\%\\
eLIAN-20-5&43.3\%&46.7\%&60.5\%&56.6\%&73\%&76.7\%& 34.5\%& 51.6\%& 50.5\%\\
\hline
\end{tabular}}
\end{table}

Success rates of the algorithms are reported in \tablename \ \ref{tab1}. As one can see eLIAN always solves more tasks that the predecessor, no matter which collection or angle constraint is used. The most notable difference is predictably between LIAN-20 and eLIAN-20-5. Suggested technique of dynamic $\Delta$-value adjusting leads to solving up to 20\% more tasks (see Baldur's Gate-$20^\circ$) and thus substantially increases the effectiveness of the algorithm. The normalized number of instances that were not solved by LIAN but solved by eLIAN is presented in \tablename \ \ref{tab2}. In absolute numbers LIAN failed to solve 659 tasks for $\Delta=20^\circ$, 451 - for $\Delta=25^\circ$ and 299 - for $\Delta=30^\circ$. Analyzing the figures, one can claim that on average eLIAN is able to successfully handle (within 5 minutes time cap) half of the tasks that LIAN is unable to solve.

\begin{figure}
\centering
\includegraphics[width=0.99\textwidth]{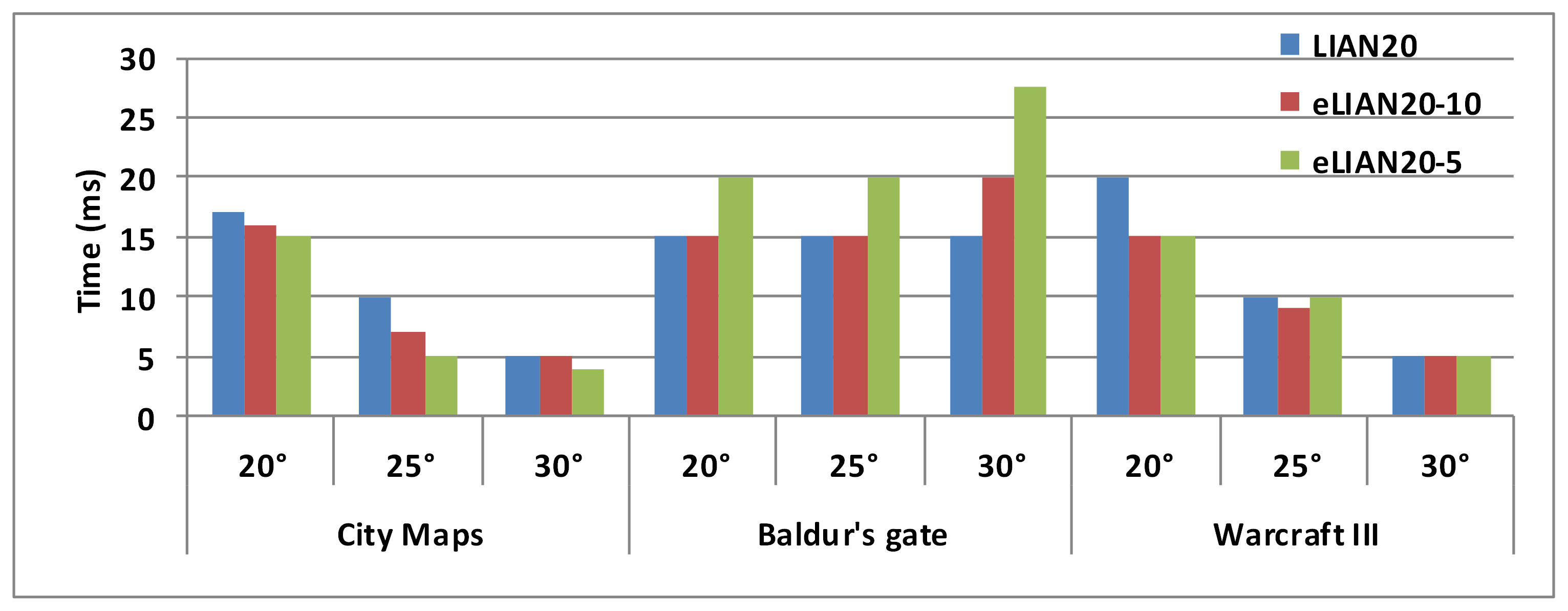}
\caption{Averaged runtime of LIAN and eLIAN}
\label{fig3}
\end{figure}

To conduct a fair comparison of the efficiency indicators, e.g. the resultant runtimes, we averaged those values across the intstances that were solved by all three algorithms: LIAN-20, eLIAN-20-10 and eLIAN-20-5. Instead of arithmetic mean value, we report the median, since there is a big discrepancy in the algorithms' runtimes (up to one order of magnitude). The results are depicted on \figurename \ \ref{fig3}. As one might see on City Maps and Warcraft III collections eLIAN works faster than the predecessor, while on Baldur's Gate eLIAN's runtime is equal or higher than the one of LIAN. This might be explained by the difference in types of environments. City Maps and Warcraft III collections consist of the maps representing outdoor environments populated with stand-alone and rather small obstacles, while most maps in Baldur's Gate collection represent indoor environments with rooms, corridors and walls. Due to the increased branching factor (e.g. re-expnaing the nodes with attempting different $Delta$-values) in case of getting stuck into obstacles represented by walls, eLIAN has to spend more time circumnavigating them in comparison with LIAN.

\begin{figure}
\centering
\includegraphics[width=0.99\textwidth]{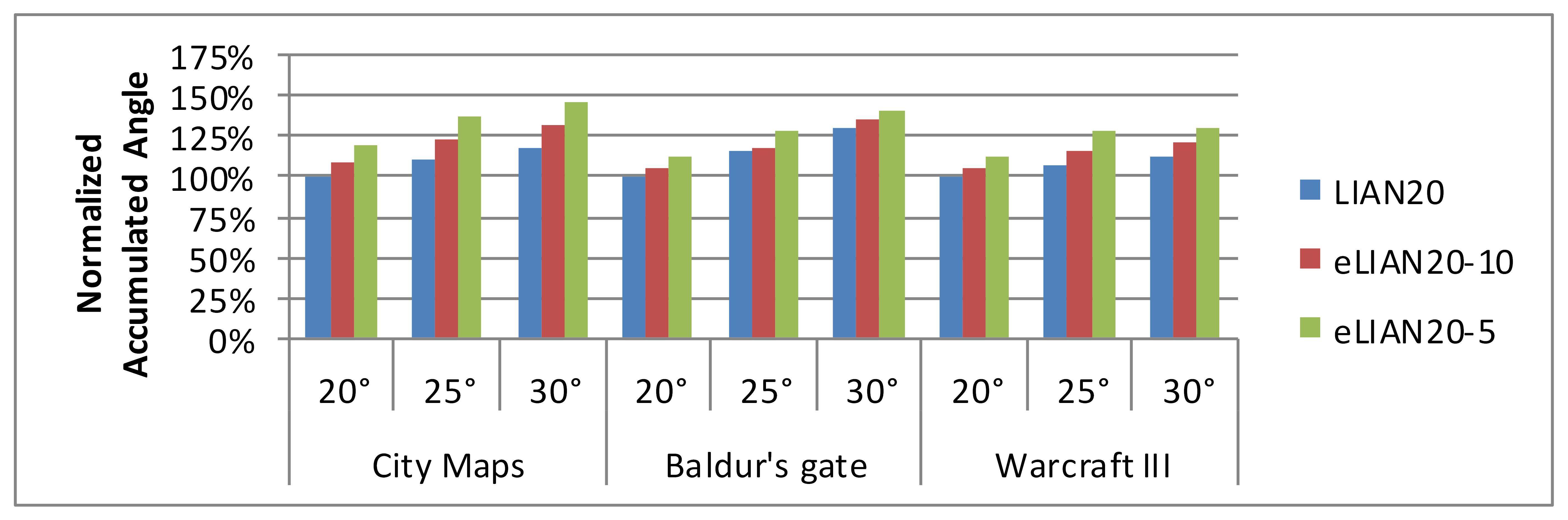}
\caption{Normalized accumulated turning angle}
\label{fig4}
\end{figure}

We have also compared the quality indicators of the solutions, provided by the algorithms. In terms of pathlength, all algorithms with all values of $\Delta$ showed very close results, and the difference between them doesn't exceed a couple of percent. Another quality indicator we were interested in represents the accumulated turning angle that the resulting trajectories contain. The normalized results are depicted on  \figurename \ \ref{fig4}. As a baseline we have taken LIAN with  20$^\circ$- angle constraint. The results on all collections show the same trends: increasing the maximum possible turning angle leads to higher values of accumulated angle. Moreover, eLIAN with allowance of making sections with less size finds trajectories with higher values of this indicator as well. This result can be explained by the fact, that the cost function of LIAN and eLIAN is targeted just on minimizing the path length and doesn't take into account the considered indicator.  As a result, the algorithm that has more opportunities to explore the search-space, e.g. eLIAN, finds paths that contain more turns.

\section{Conclusions}
We have considered a problem of grid-based angle-constrained path planning and presented a novel technique that significantly boosts the performance of the state-of-the-art planner tailored to solve those type of problems. As shown experimentally, success rate of the enhanced algorithm is notably higher (up to 20\%) while solution quality is nearly the same. Obvious and appealing direction of future research is elaborating on further enhancements to make the algorithm complete. 

\section{Acknowledgments}
The work was partially supported by the ``RUDN University Program 5-100'' and by the special program of the presidium of Russian Academy of Sciences.

\bibliographystyle{splncs04}
\bibliography{cites}

\end{document}